\documentclass{article}
\pdfoutput=1
\usepackage{pdfpages}
\usepackage{booktabs}
\usepackage{algorithmic}
\usepackage{graphics}
\usepackage{cite}
\usepackage{amsmath} 
\usepackage{multicol}
\usepackage{url}
\usepackage{latexsym}

\title{Online Deep Learning: Growing RBM on the fly}
\author{Savitha Ramasamy, Kanagasabai Rajaraman, \\
	Pavitra Krishnaswamy and Vijay Chandrasekhar \\
	Institute for Infocomm Research, A*STAR, Singapore  \\
	Email: \{ramasamysa, kanagasa, pavitrak, vijay\}@i2r.a-star.edu.sg\\}

\date{\today} 

\begin{document}
\maketitle
\begin{abstract}
  We propose a novel online learning algorithm for Restricted Boltzmann Machines (RBM), namely, the Online Generative Discriminative Restricted Boltzmann Machine (OGD-RBM), that provides the ability to build and adapt the network architecture of RBM according to the statistics of streaming data. The OGD-RBM is trained in two phases: (1) an online generative phase for unsupervised feature representation at the hidden layer and (2) a discriminative phase for classification. The online generative training begins with zero neurons in the hidden layer, adds and updates the neurons to adapt to statistics of streaming data in a single pass unsupervised manner, resulting in a feature representation best suited to the data. The discriminative phase is based on stochastic gradient descent and associates the represented features to the class labels. We demonstrate the OGD-RBM on a set of multi-category and binary classification problems for data sets having varying degrees of class-imbalance. We first apply the OGD-RBM algorithm on the multi-class MNIST dataset to characterize the network evolution. We demonstrate that the online generative phase converges to a stable, concise network architecture, wherein individual neurons are inherently discriminative to the class labels despite unsupervised training. We then benchmark OGD-RBM performance to other machine learning, neural network and ClassRBM techniques for credit scoring applications using 3 public non-stationary two-class credit datasets with varying degrees of class-imbalance. We report that OGD-RBM improves accuracy by 2.5-3\% over batch learning techniques while requiring at least 24\%-70\% fewer neurons and fewer training samples. This online generative training approach can be extended greedily to multiple layers for training Deep Belief Networks in non-stationary data mining applications without the need for \emph{a priori} fixed architectures.
\end{abstract}

\section{Introduction}
Deep learning algorithms have superlative capabilities for jointly performing feature mapping and classification. Thus, they outperform other machine learning approaches in applications ranging from image classification \cite{Article_alex} and medical diagnostics \cite{Article_Turner} to credit fraud analytics \cite{Article_classRBM3}. However, it is challenging to adaptively (re)train deep neural networks to track changes in data distribution, especially in streaming data applications. Moreover, training multiple layer neural networks requires \emph{a priori} specification of a suitable network architecture, and thus it is difficult to inform choice of architecture with the statistics of the data. 

Online learning approaches for deep neural networks have the potential to address both these challenges. Several studies have put forth online learning algorithms for training single layer perceptron networks \cite{Article_Online1,Article_Onlinernn,Book_online}. Single layer feedforward neural networks can be trained in an online fashion using Stochastic Gradient Descent \cite{Article_Online2} or Extended Kalman Filters \cite{Article_mccit2fis}\cite{Article_csran} for the parameter update. However, it remains challenging to extend these successes to the task of training deep neural networks in a fully online manner. For example, online algorithms for denoising autoencoders (DAE) \cite{Article_onlineAE} have been used for incremental feature learning with streaming data, but need \emph{a priori} training with a DAE architecture as the building block to learn a base set of features first. Further, incremental learning has been applied within a boosting convolutional neural network framework for feature augmentation, loss function updation and fine-tuned back propagation with information accumulating in successive mini-batches \cite{Article_ocnn_nips2016}.  Finally, it has also been shown that updating a greedily pre-trained layer-wise restricted Boltzmann machines (RBMs) in an online fashion automatically learns discriminative features for classification \cite{Article_OnlineRBM}. However, all the above approaches require pre-training and/or a fixed base network architecture as a precursor for incremental online updates with streaming data.  Thus, methods that evolve a network architecture from scratch in an online manner, as the data streams in, would offer novel capabilities for online learning with deep neural networks.

In this paper, we present an unsupervised online learning algorithm named the Online Generative Discriminative Restricted Boltzmann Machine (OGD-RBM), for generative RBM and hence, layer-wise training of DBN. At the beginning, there are no neurons in the hidden layer of a RBM. As samples stream in, the ability of the network to represent the sample is assessed using the reconstruction error for the sample. Based on this reconstruction error, the algorithm either deletes the samples that are well represented, or adds a neuron to the hidden layer to represent the sample or updates the weights for existing neurons in the network. As the network updates are tailored to represent the distributions of the distinctive input features, the network is compact and inherently discriminative \cite{Article_DDFA2015}. Finally, the features learned in the generative phase are mapped to the specific classes via discriminative learning.

We first demonstrate the unique abilities of the OGD-RBM to represent the distinctive class distributions and to learn in a manner that is invariant to the training data sequence, through a study on the well-explored MNIST data set. The sequential invariance is much like the invariance to permutations in the training set seen with batch learning algorithms \cite{Article_onlineStab2011}. We then evaluate the performance of OGD-RBM on binary classification tasks with a variety of highly unbalanced streaming credit fraud analytics datasets. It is critical to learn the distribution of a minority class from a highly imbalanced data set, and online learning provides a premise to efficiently learn the under-represented minority class, owing to its ability to detect novelty in data. Our results show that the OGD-RBM can perform better than batch DBN with lesser network resources and fewer training samples than batch methods. The main contributions of the paper are:
\begin{itemize}
	\item An online generative learning algorithm for unsupervised feature representation at the hidden layer of RBM.
	\item A statistical demonstration that the neurons trained through the unsupervised online generative training are inherently discriminative.
	\item A statistical demonstration that the classification accuracy and the neuron-to-class label associations of the OGD-RBM are independent of the sequence in which the training samples are presented.
	\item A demonstration that the OGD-RBM achieves better accuracies with more compact network architectures than batch learning algorithms.
\end{itemize}

The paper is organized as follows. First, we present the OGD-RBM architecture and algorithm. Next, we demonstrate the learning algorithm of OGD-RBM using the MNIST. Then, we evaluate OGD-RBM in relation to other algorithms applied to the credit fraud detection problem. Finally, we summarize the study and outline future directions.

\section{Online Generative Discriminative Restricted Boltzmann Machine}\label{OL-McRBM}
\begin{figure}
	\includegraphics[width=0.5\textwidth]{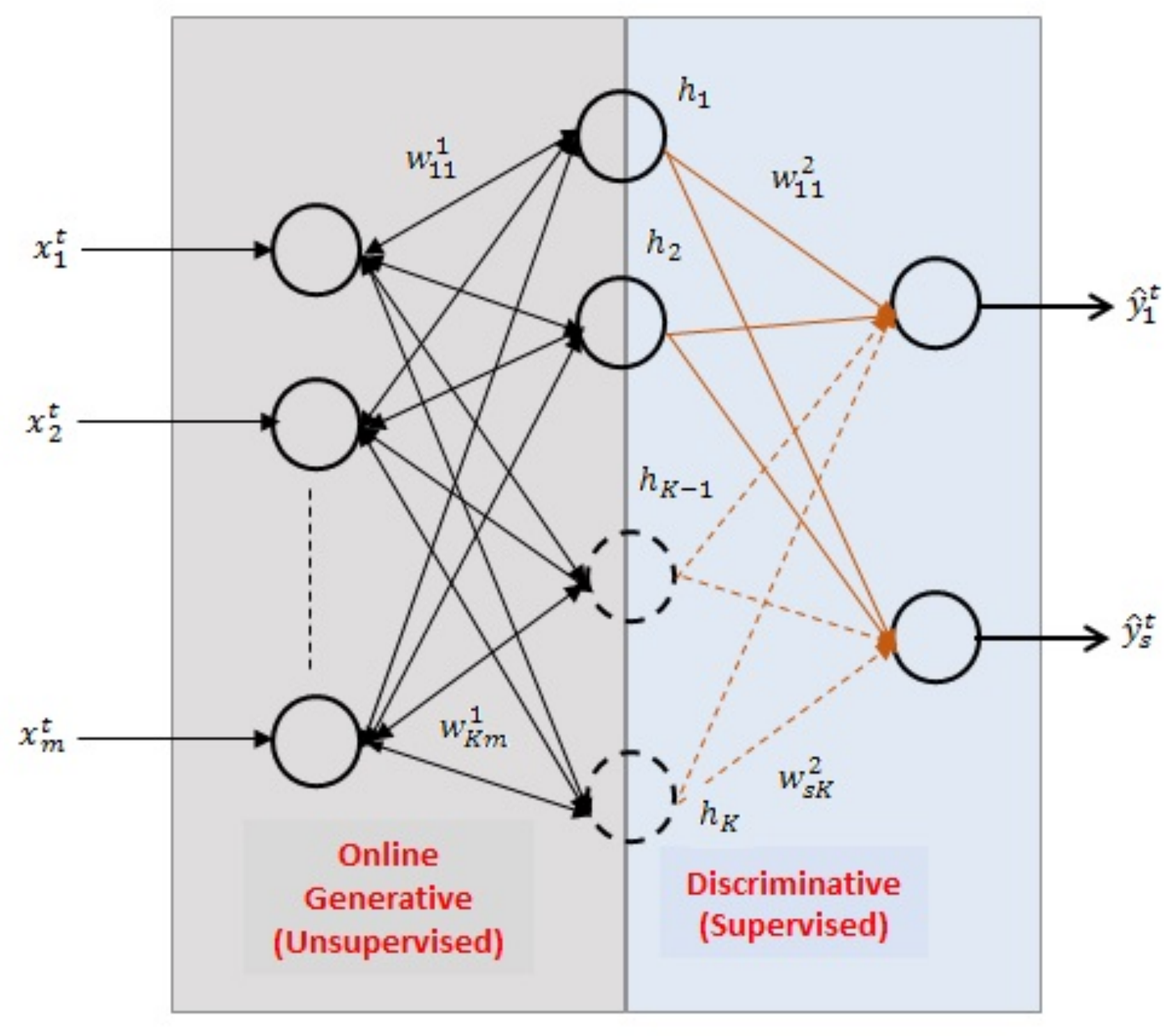}\caption{The network architecture and the training phases. Online generative learning is performed in the first phase, wherein the network begins with zero neurons in the hidden layer, and adds and/or adapts the network learning progresses to derive a feature representation of the data in an unsupervised manner. The next phase performs supervised discriminative modeling to associate the feature representation with the class labels.}\label{nwstruct}
\end{figure}

We describe the Online Generative Discriminative Restricted Boltzmann Machine (OGD-RBM) learning algorithm. Fig. \ref{nwstruct}  shows the two phases of training the OGD-RBM, namely: (1) An online generative learning phase for unsupervised feature representation at the hidden layer, and (2) a discriminative phase for supervised modeling of the class conditional probabilities.

We denote the training data set as $\left\{\left(\mathbf{x}^{1},c^1\right),\ldots,\left(\mathbf{x}^{t},c^t\right),\ldots\left(\mathbf{x}^{N},c^N\right)\right\}$, wherein $\mathbf{x}^{t}\in \Re^{m} = [x^{t}_1, \ldots, x^{t}_j,\ldots,x^{t}_{m}]$ is the $m$-dimensional input of the $t^{\rm{th}}$ sample; $c^t \in \{1,2,\ldots,s\}$ denotes the set of class labels or targets for $s$ classes, and $N$ is the total number of samples. The objective of the OGD-RBM is to best approximate the functional relationship between the inputs and their targets.

\subsection{Preliminaries}
A Restricted Boltzmann Machine (RBM) \cite{Article_HIN02} has visible and hidden layers, connected through symmetric weights. The inputs ($\mathbf{x} =[x_1 \ldots x_m]^T$) correspond to the neurons in the visible layer. The response of the $K$ neurons in the hidden layer ($\mathbf{h} = [h_1 \ldots h_K]^T$) model the probability distribution of the inputs. The probability distribution is derived by learning the symmetrical connecting weights between the visible and the hidden layers ($w_{ji}^1;~ j =1,\ldots,m,~ i = 1,\ldots,K$). The neurons in the same layer of the RBM are not connected.

The conditional probability of a configuration of the hidden neurons ($\mathbf{h}$), given a configuration of the visible neurons associated with inputs $\mathbf{x}$, is:
\begin{eqnarray}
P(\mathbf{h}|\mathbf{x})&=&\prod_{i=1}^K P(h_i|\mathbf{x}).
\end{eqnarray}

The objective of the training phase is to learn the unknown ($\mathbf{h}$) iteratively using the input ($\mathbf{x}$), as described below. Denote a configuration of the visible neurons by ($\mathbf{\widehat{x}}$). Then, the conditional probability of ($\mathbf{\widehat{x}}$) given a configuration of the hidden neurons ($\mathbf{h}$), is:
\begin{eqnarray}
P(\mathbf{\widehat{x}}|\mathbf{h})&=&\prod_{j=1}^m P(\widehat{x}_j|\mathbf{h}).
\end{eqnarray}
Denote a configuration of the hidden neurons by ($\mathbf{\widehat{h}}$). Then, the conditional probability of ($\mathbf{\widehat{h}}$) given ($\mathbf{\widehat{x}}$) is:
\begin{eqnarray}
P(\mathbf{\widehat{h}}|\mathbf{\widehat{x}})&=&\prod_{i=1}^K P({\widehat{h}_i}|\mathbf{\widehat{x}}).
\end{eqnarray}
The individual activation probabilities are given by:
\begin{eqnarray}
P(h_i=1|\mathbf{x})&=&\sigma\left(b_i+\sum_{j=1}^{m}w_{ji}^1 x_j\right)\label{hidpact},\\
P(\widehat{x}_j=1|\mathbf{h})&=&\sigma\left(c_j+\sum_{i=1}^{K}w_{ji}^1 h_i\right)\label{inpact},\\
P(\widehat{h}_i=1|\mathbf{\widehat{x}})&=&\sigma\left(b_i+\sum_{j=1}^{m}w_{ji}^1 \widehat{x}_j\right),\label{hidpact1}
\end{eqnarray}
where the function $\sigma(.)$ refers to the sigmoidal activation function. The generative training phase iterates until the ($\mathbf{\widehat{x}}^t$) most closely approximates ($\mathbf{x}^t$).

Training is performed using the maximum likelihood criterion, implemented by minimizing the negative log probability of the training data $L_{gen}$:
\begin{equation}
L_{gen} = - \sum_{x\in N} log~ P(\mathbf{x}|(w_{ji},b_i,c_j)). \label{mlgen}
\end{equation}

The weights between the input and hidden layers of the RBM are updated according to:
\begin{equation}
w_{ji}^1 = \alpha*(x_j*h_i-\widehat{x}_j*\widehat{h}_i),\label{weightup}
\end{equation}
wherein $\alpha$ denotes a pre-specified learning rate. 

\subsection{Online Generative Learning}\label{Gme}
We now describe the online generative learning process for feature representation at the hidden layer. Initially, the hidden layer has no neurons. As the data streams in, the online generative learning algorithm of the RBM adds neurons (with activations defined by Eqs. \eqref{hidpact} and \eqref{inpact}) and/or updates the representations of the existing neurons depending on the novelty of the sample. The first neuron is added based on the first sample in the data set.

At a given point during the training process, the network comprises $K-1$ neurons in the hidden layer, corresponding to $K-1$ novel samples out of a history of $t-1$ samples presented to the RBM thus far. For the next $t^{\rm{th}}$ sample, the reconstruction error of the network is: 
\begin{equation}
E_{cd}^t = \frac{1}{m} \sum_{i=1}^m (x_i^t-\widehat{x}_i^t)^2.
\end{equation}
The reconstruction error ($E_{cd}^t$) is compared to two pre-defined thresholds, namely the novelty threshold $E_n$ and the marginal representation threshold $E_m$. Based on this comparison, the algorithm chooses one of the following steps for the $t^{\rm{th}}$ sample:
\begin{itemize}
	\item {\bf Add a Representative Neuron:} If $(E_{cd}^t > E_n$), the sample is deemed novel and a neuron ($K$) is added to the hidden layer of the network. The input weights connecting the $K^{\rm{th}}$ hidden neuron and the neurons in the input layer are obtained as a function of the inputs $g\left(\mathbf{x}^t\right)$, where $g\left(.\right)$ can be any function such that $\mathbf{w}_i^1 = [w_1i \cdots w_mi]^T \in\Re^m\in(0,1)$. In this work, we assign $\mathbf{w}_i^1 = 0.01*\mathbf{x}^t$. The network weights of all the neurons, including the new neuron, are then updated according to Eq. \eqref{weightup}.
	\item {\bf Adapt Existing Network:} If $E_n>E_{cd}^t>E_m$, the network weights ($w_{ji}^1, i=1,\cdots,K, j = 1,\cdots,m$) are adapted such that the probability distribution approximated by the hidden neurons includes the representation of this sample, according to Eq. \eqref{weightup}.
	\item {\bf Ignore Sample:} If $E_{cd}^t < E_m$), then the sample is sufficiently represented by the existing network and does not warrant a network update.
\end{itemize}
Overall, the neurons in the hidden layer of the network are adaptively added and updated to obtain a compact network structure that is sufficiently representative of the data. 

\subsection{Discriminative Training}
We now describe the discriminative training, where the feature representation learned during the online generative phase is mapped to the conditional class distributions in a supervised fashion. 

The responses of the $K$ neurons in the hidden layer are as below:
\begin{eqnarray}
\mathbf{h} = \left[h_1 \ldots h_K\right]^T
\end{eqnarray}
This feature representation is then used in a supervised discriminative training phase to learn the conditional probability distribution $P(c^t|\mathbf{x}^t)$. The class labels $c^t$ are encoded in $\mathbf{y}^t = [y_1^t,~\cdots,~y_s^t]$, as below:
 \begin{equation}
 y_i^t = \left\{\begin{array}{cc}
 1 & \text{if}~~ c^t=i, \\
 0 & \text{otherwise}. \\
 \end{array} \right.i=1,\ldots s;
 \end{equation}

The objective of discriminative training is to minimize the log probability
\begin{eqnarray}
\min_{\mathbf{w}_{ki}^2} \frac{1}{N} \sum_{n\in N} \mathcal{L}_{disc} \left(\mathbf{y}^n|\mathbf{x}^n\right), \label{outputwt}
\end{eqnarray}
where $\mathcal{L}_{disc} \left(\mathbf{y}^n|\mathbf{x}^n\right)$ is a measure of error between $\mathbf{y}^n$ and $\mathbf{\widehat{y}^n}$, and $\mathbf{w}_{ki}^2$ are the weights connecting the $k$-th output neuron and the $i$-th hidden neuron. Here, we perform discriminative training through 10 epochs of supervised training using a Multi-Layer Perceptron (MLP) with sigmoidal activation function.

\section{Demonstration of OGD-RBM: MNIST}\label{mnist}
We now demonstrate the progression of learning within the proposed OGD-RBM approach, and make some observations about the algorithm. We characterize the algorithm on the MNIST data set \cite{Article_MNIST}, as it is a large well-explored multi-category dataset (60,000 training samples, 10 categories) to demonstrate the OGD-RBM. The network is trained in an online fashion, using the training data set. The validity of the trained network is established independently on the test set (10,000 samples, 10 categories) in an offline fashion.
\begin{figure*}[h]
	\begin{center}
		\includegraphics[width=\textwidth]{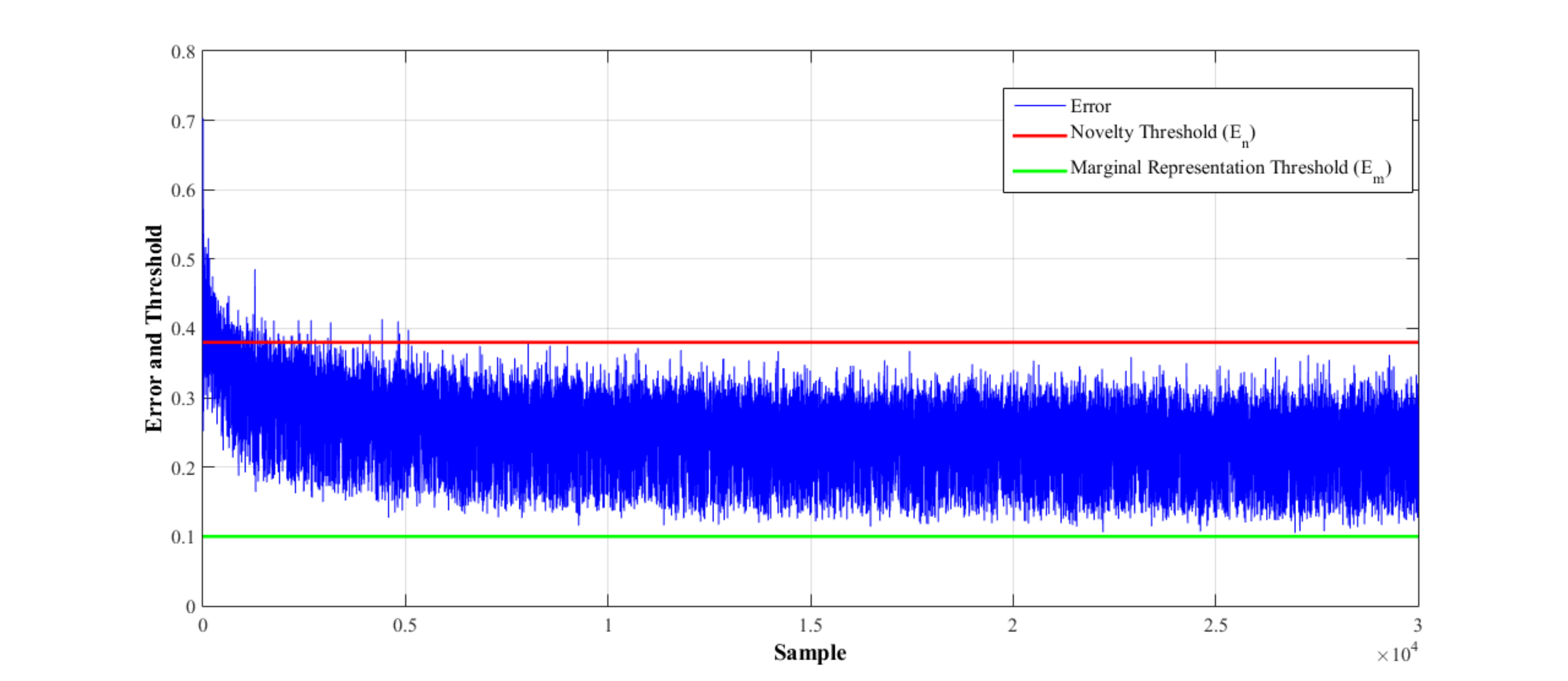}\caption{Reconstruction error of the network for samples numbered 1 through 30000, shown in relation to the learning thresholds. The error stabilizes after about $5000$ samples.}\label{Error_evolve}
	\end{center}
\end{figure*}
\begin{figure*}[h]
	\begin{center}
			\includegraphics[width=\textwidth]{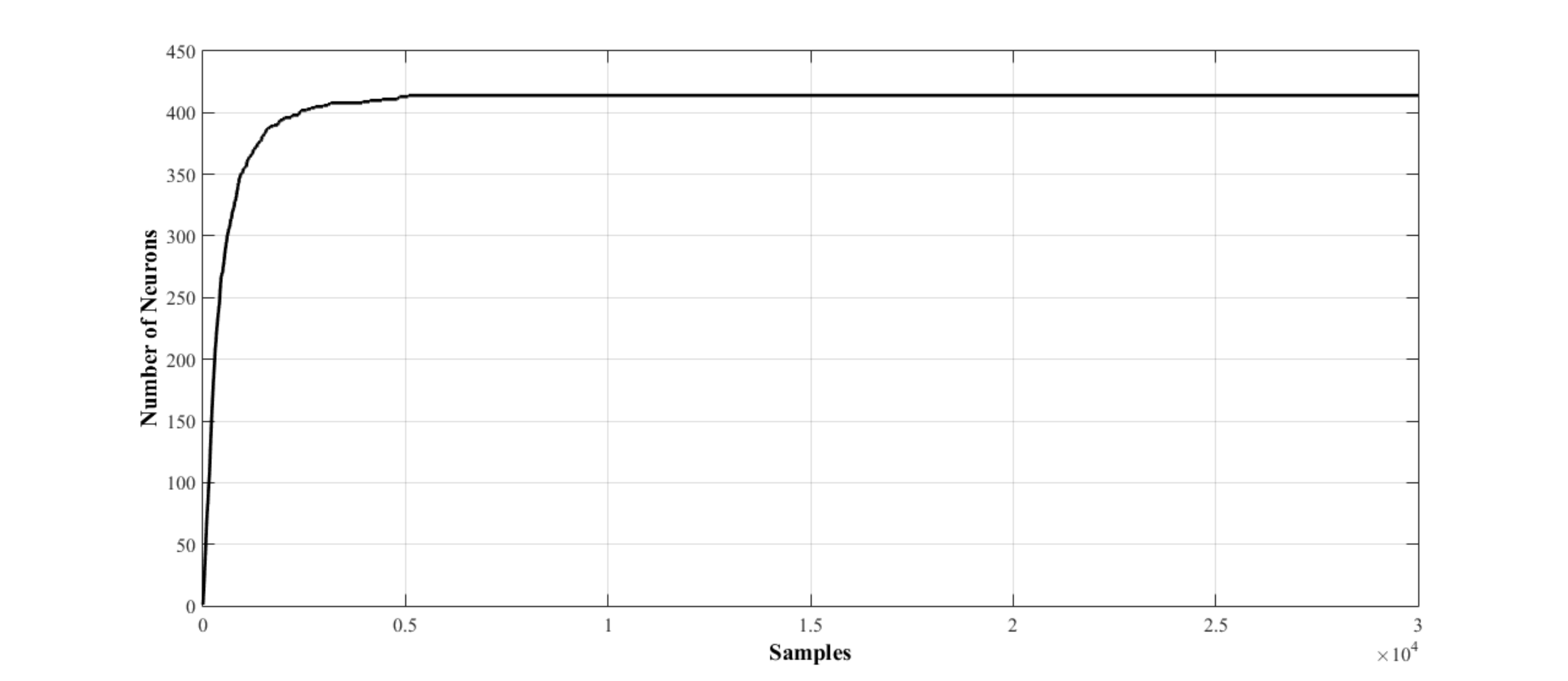}\caption{Number of neurons in the network as training proceeds from sample number 1 through sample number 30000}\label{Neuron_growth}
	\end{center}
\end{figure*}

Fig. \ref{Error_evolve} and \ref{Neuron_growth} show the evolution of reconstruction error and network architecture, as samples stream in for training. Fig. \ref{Error_evolve} shows that the reconstruction error is high for the initial samples. This is because the model is at infancy and is beginning to learn. Hence, most samples are novel to the network, resulting in neurons being added (see Fig. \ref{Neuron_growth}). However, as training progresses, the network learns a sufficient representation of the data and the reconstruction error reduces progressively, resulting in fewer neurons being added to the network. It is evident from Fig. \ref{Error_evolve} and \ref{Neuron_growth} that the online generative phase converges to a stable, concise network architecture, and the generative training is complete in 26min $\pm$ 15. It is also evident from Fig. \ref{Neuron_growth} that $~90$\% of the neurons in the stable network are added for the first $10$\% of the training samples (i.e., the first $~5000$ samples). The remaining $90$\% of the training samples (i.e., the latter $~55000$ samples) contribute only about $10$\% of the neurons in the stable network.

We next tested the effect of the training data sequence on the performance of the algorithm. We trained the OGD-RBM independently for $50$ randomly constructed sequences of the MNIST training samples. In each case, we presented different sequences of the training data set to train the network. Across the $50$ training trials, the classification accuracy on the testing data set is 97\%pm 2\%, and the final number of neurons was $403\pm 26$ . Thus, changing the sequence of presentation of training samples does not change the accuracy or the network architecture significantly, showing that the network is able to generalize well with a concise network architecture. 

\begin{figure*}[h]
	\begin{center}
		\includegraphics[width=\textwidth]{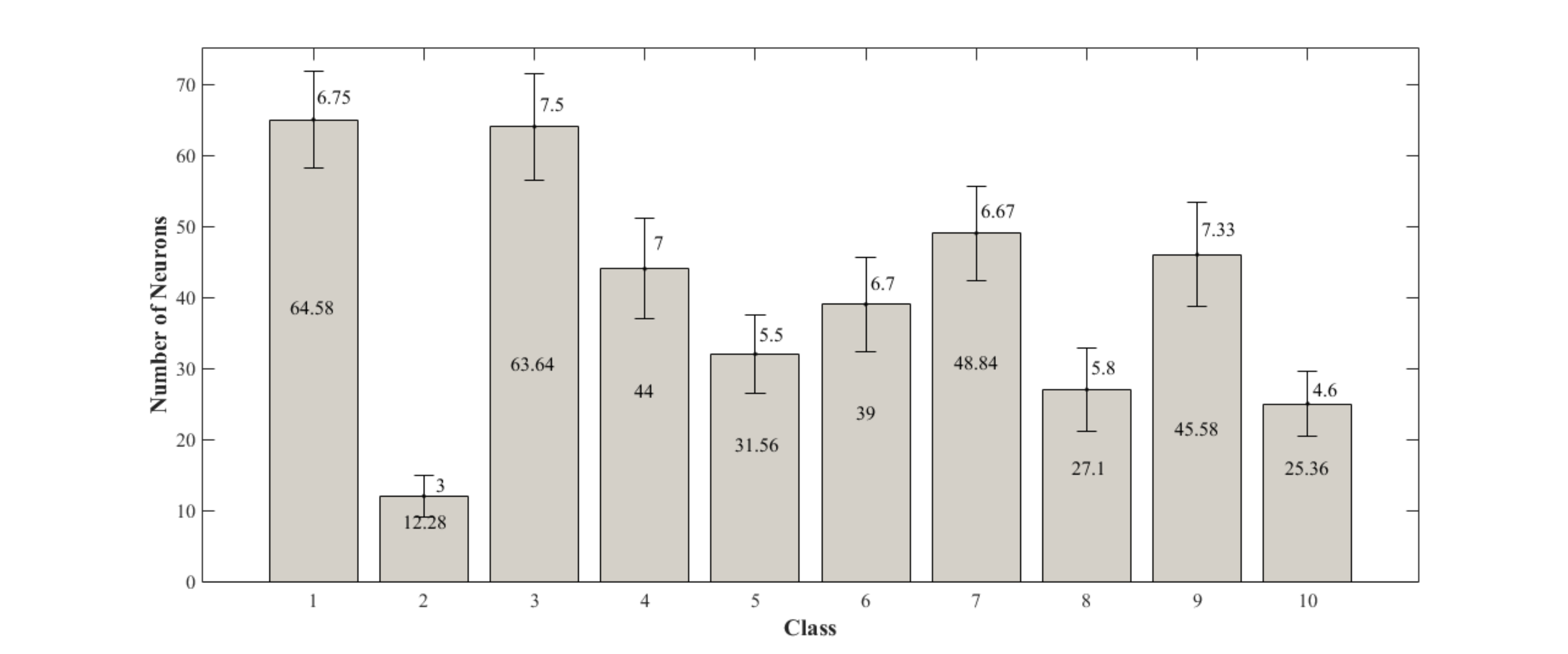}\caption{MNIST Classification: Average number of hidden layer neurons associated with each class of the MNIST dataset, with the standard deviation across $50$ trials. 
			}\label{neurondist}
	\end{center}
\end{figure*}

To study the discriminative potential of the feature representation learned during the online generative training phase, we related the number of `novel' samples (where $E_{cd}^t > E_n$) to their corresponding class labels $c^t$ for each of the $50$ trials. Fig. \ref{neurondist} shows the average number of hidden layer neurons associated with each class of the MNIST dataset, with the standard deviation across the $50$ trials. These results show that the individual neurons in the trained network are inherently discriminative to the class labels despite the unsupervised nature of the training. Further, we observe that the variability across trials is a small proportion of the average number of neurons in each class, suggesting that the neuron-to-class associations are largely independent of the sequence of training data samples.

\section{Performance Study: Credit Fraud Analytics}\label{credit-fraud}
Online learning algorithms are particularly suitable for streaming data applications where the data distribution evolves with time, and are therefore relevant to the problem of credit scoring. Credit scoring is the problem of estimating the probability that borrower might default and/or exhibit undesirable behavior in the future. This problem is usually characterized with an imbalanced data set, and there is a huge inter-personal variability across borrowers. Such a problem calls for online learning algorithms that is capable of learning the distribution of the data set, regardless of the imbalance and distinction between samples in the same class.

Several studies have employed batch machine learning techniques for credit scoring (\cite{Article_credit3}, \cite{Article_credit1}, \cite{Article_credit9}, \cite{Article_classRBM3}). We perform analogous evaluations to benchmark our online learning algorithm in relation to these batch learning techniques. Specifically, we perform credit fraud prediction using three publicly available data sets, namely, the UCI German credit data set (UCI German), the UCI Australian credit data set (UCI AUS), and the KAGGLE 'Give me some credit' data set (KAGGLE GMSC). We evaluate the OGD-RBM classifier, in comparison with the Support Vector Machine classifier (SVM), the Multi-layer Perceptron Neural Network (NN) classifier, the Classification Restricted Boltzmann Machine classifier (ClassRBM) \cite{Article_classRBM2} and the Scoring Table (ST) method on the three credit data sets (listed in Table \ref{creditdata}) in Table \ref{cred17030}.

Table \ref{creditdata} details the public credit scoring data sets, along with the the number of classes, the number of training and testing samples, and their imbalance factors ($I.F.$):
\begin{equation}
{I. F.} = 1- \frac{s}{N}~\min_{i=1\cdots s}~N_i, \label{IFAC}
\end{equation}
where $s$ is the total number of classes and $N_i$ is the number of samples in class $i$. It is evident that the 3 public datasets have varying degrees of class-imbalance. While the UCI AUS is mildly imbalanced, the UCI German is partially imbalanced  and the KAGGLE GMSC has very high imbalance across classes. This varying degree of class-imbalance provides a unique opportunity to characterize the neuron distribution across classes in the online learning framework. 
\begin{table}[htp]
	\begin{center}
		\caption{Description of the credit scoring data sets}\label{creditdata}
		\begin{tabular}{|l|c|c|c|c|}\hline\hline
			Data set         & Input    & Size of         & $I. F.$   \\
							 & features & Data set        &           \\\hline
			UCI AUS          &   14     &    690          & 0.1101    \\
			UCI German      &   24     &    1000         & 0.4       \\
			KAGGLE GMSC      &   10     &    150000       & 0.86632   \\\hline\hline
		\end{tabular}
	\end{center}
\end{table}
We filled in the missing values in the Kaggle 'Give me some credit' data set by averaging across similar participants in the population, grouped according to ages in intervals of $10$.

We compare the classifiers on the three problems based on the network size and the performance measures such as the overall efficiency ($\eta_O$), the average efficiency ($\eta_A$), True Positive Rate (TPR), True Negative Rate (TNR), and Geometric mean accuracy (Gmean) defined as:
\begin{eqnarray}
\eta_O &=& \frac{\sum_{i=1}^s q_{ii}}{N}\text{X}100\%\label{etao}\\
\eta_A &=& \frac{1}{s}\sum_{i=1}^s \frac{q_{ii}}{N_i}\text{{X}}100\%\label{etaa}\\
\text{TPR} &=&\frac{\text{Number of TP}}{\text{Number of TP}+\text{Number of FN}}\label{TPR}\\
\text{TNR} &=& \frac{\text{Number of TN}}{\text{Number of TN}+\text{Number of FP}}\label{TNR}\\
\text{Gmean}&=&\sqrt{\text{TPR}\times\text{TNR}}\label{Gme}
\end{eqnarray}
Here, $q_{ii}$ is the the number of correctly classified samples in class $i$ and $N_i$ is the total number of samples in class $i$.

We now present the results of OGD-RBM in relation to the batch learning techniques. We reproduce previously obtained batch learning results using the SVM, NN, ClassRBM and ST classifiers from \cite{Article_classRBM3}. Although the ClassRBM results in \cite{Article_classRBM3} are reported with fixed architecture of $100$ neurons with a batch size of $100$ and learning rate of $0.0001$, the architecture of the other classifiers is not specified. Further, the training accuracies of the classifiers have also not been reported. Hence, we perform independent evaluations using SVM, NN, and ClassRBM classifiers, to report an additional performance validation beyond the previously reported results.

The performance comparisons provide the following observations:
\begin{itemize}
\item {\bf Network Size:} Overall, the OGD-RBM network uses fewer neurons than the classifiers used in comparison. This is because the OGD-RBM uses the most novel samples to add neurons to the network, and the neurons are well-representative of the data set.
\item {\bf Performance Measures:} Despite having a compact architecture, the proposed OGD-RBM performs better than all the classifiers used in comparison. This could be attributed to the fact that the learnt distributions represent the data very well. Moreover, while the other algorithms learn the data in batches, and updates gradients in batches, the OGD-RBM updates gradients based on every sample in the data set.
\item {\bf Neuron Distribution Per Class:} Unlike the batch learning algorithms that need {\it a priori} assumption of the architecture, the OGD-RBM builds the network as learning progresses. This helps us to infer the number of neurons per class that may help to characterize the distribution of the samples in each class.
\item {\bf Effect of Class-Imbalance:} Classes with fewer samples require more neurons for sufficient feature representation. As the class imbalance increases, a greater proportion of the hidden layer neurons is associated with less prevalent classes. This adaptation is a natural consequence of the online learning process, and differentiates our approach from the the batch learning algorithms. 
\end{itemize}

\begin{table*}[htp]
	\caption{Performance results on the credit scoring data sets: Overall and Average Efficiencies} \label{cred17030}
	\begin{center}
		\begin{tabular}{|c|c|c|c|c|c|c|c|c|c|c|}\hline\hline
			Data set        & Classifier&   $K$  &   \multicolumn{2}{|c|}{Training} & \multicolumn{5}{|c|}{Testing}\\\cline{4-10}
							&           &        &  $\eta_O$   & $\eta_A$      & $\eta_O$  & $\eta_A$ & TPR    & TNR     & Gmean \\\hline
			                &  SVM      & 534    &   76.429    & 66.679         & 74.667   & 61.378   & 0.3255 & 0.8878  &  0.5376\\\cline{2-10}
			                &  SVM*     & -    &  -          &   -            & -        &  -       & 0.484 & 0.867  &  0.648\\\cline{2-10}
			UCI      	&  NN       &  60    &  98.571     & 97.573         & 72.333   & 65.105   & 0.4574 & 0.8446  &  0.6216\\\cline{2-10}
							&  NN*       &  -   & -           & -              & -   & -   & 0.517 & 0.814 &  0.648\\\cline{2-10}
			German         &  ClassRBM &  80    &  77.428     & 63.346         & 74.000   & 56.738   & 0.4418 & 0.8271  &  0.6045\\\cline{2-10}
							&  ClassRBM* & 100   & -    & -        & -   & -   & 0.479& 0.872  &  0.646\\\cline{2-10}
			         &  ST*      & -   & -    & -        & -   & -   & 0.67& 0.68  &  0.68\\\cline{2-10}
			&  \textbf{OGD-RBM}   & \textbf{48 }& \textbf{79} & \textbf{74.2} &  \textbf{76.5}  &  \textbf{71.69}&\textbf{0.60} & \textbf{0.83} & \textbf{0.71} \\
			&&(32:16)&  &  &    &  & &  & \\\hline			
		                  	&  SVM      & 192    &         85.507  & 86.263         & 85.507   & 86.048  & 0.7946 & 0.9263  & 0.8579\\\cline{2-10}
		                  	&  SVM*     & -    &        -  & -         & -   & -  & 0.913 & 0.71  & 0.850\\\cline{2-10}
			UCI      &  NN       &  60    &        94.824   & 94.767         & 84.058   & 83.727  & 0.7917 & 0.8828  & 0.836 \\\cline{2-10}
							&  NN*      & -    &       -   & -     & -   & -  & 0.850 & 0.857  & 0.854 \\\cline{2-10}
			AUS     		&  ClassRBM &  50    &        86.128   & 86.391         & 85.507   & 86.021  & 0.8953 & 0.8264  & 0.8602\\\cline{2-10}
			                &  ClassRBM*& 100    &        -   & -   &- & -& 0.880 & 0.847  & 0.863\\\cline{2-10}
			        &  ST*      & -   &        -   & -   &- & -& 0.828 & 0.805  & 0.816\\\cline{2-10}
			&  \textbf{OGD-RBM}   & \textbf{38 }& \textbf{86.68} & \textbf{86.8} &  \textbf{88.49}  &  \textbf{89}&\textbf{0.92} & \textbf{0.86} & \textbf{0.89} \\
			&&(20:18)&&  &    &  & &  & \\\hline					
			                &  SVM      &  6340  &        69.970   & 59.430         & 72.240   & 60.018  & 0.5771 & 0.8982  & 0.72  \\\cline{2-10}
			                &  SVM*     &  -     &       -   & -   & -   & -  & 0.114 & 0.994  & 0.336  \\\cline{2-10}
			KAGGLE  		&  NN       &   60   &        63.896   & 62.287         & 74.200   & 63.017  & 0.6165 & 0.8792  & 0.7363\\\cline{2-10}		
			                &  NN*      &  -     &       -  & -   & -  & - & 0.229 & 0.986  & 0.475\\\cline{2-10}		
			GMSC    &  ClassRBM &  100   &        75.687   & 74.048         & 86.160   & 74.789  & 0.6    & 0.8975  & 0.73384\\\cline{2-10}
			                &  ClassRBM*&  100   &        -   & -   & - &- & 0.182 & 0.991  & 0.424\\\cline{2-10}
			                &  ST*      &  -     &        -  & -   & - & - & 0.515 & 0.622  & 0.566\\\cline{2-10}
						&  \textbf{OGD-RBM}   & \textbf{13}& \textbf{76.08} & \textbf{74.49} &  \textbf{86.25}  &  \textbf{75.22}&\textbf{0.63} & \textbf{0.88} & \textbf{0.74} \\
						&&(3:10)&&&&&&&\\\hline\hline
		
		\end{tabular}
	\end{center}
	*Reproduced from \cite{Article_classRBM3}	
\end{table*}

\section{Conclusion}\label{conc}
We introduced a novel Online Generative Discriminative Restricted Boltzmann Machines (OGD-RBM) algorithm that evolves a network architecture in a fully bottom-up online manner as data streams in. We demonstrated that the algorithm converges to a stable compact network architecture wherein (a) hidden layer neurons are implicitly associated with class labels (despite unsupervised training), and (b) classification performance are invariant to the sequence in which the training data samples are presented. Further, OGD-RBM performed better than batch techniques in credit score classification with streaming data -- specifically online learning achieved better accuracy with fewer neurons and showed the unique ability to adapt to class imbalance. Areas of future work will include expansions to unsupervised discriminative training, deeper architectures, and interpretable models.

\end{document}